\documentclass{article}[custombib]

\usepackage[final]{neurips_2021}

\usepackage[utf8]{inputenc} 
\usepackage[T1]{fontenc}    
\usepackage{booktabs}       
\usepackage{graphicx}
\usepackage[tableposition=top]{caption}
\usepackage{chngcntr}

\usepackage{url}            
\usepackage{xurl}
\usepackage{natbib}

\title{A Recommendation System to Enhance Midwives' Capacities in Low-Income Countries}

\author{Anna Guitart \\ 
benshi.ai \\
\texttt{guitart@benshi.ai}\\
\And
Afsaneh Heydari\\
benshi.ai \\
\texttt{afsaneh@benshi.ai}\\
\And
Eniola Olaleye\\
benshi.ai \\
\texttt{eniola@benshi.ai}\\
\And
Jelena Ljubicic\\
benshi.ai \\
\texttt{jelena@benshi.ai}\\
\And
Ana Fern\'andez del R\'io\\
benshi.ai \\
\texttt{ana@benshi.ai}\\
\And
\'Africa Peri\'añez\\
benshi.ai \\
\texttt{africa@benshi.ai}\\
\And
Lauren Bellhouse \\
Maternity Foundation \\
\texttt{lauren@maternity.dk}\\
}

\begin{document}

\maketitle

\begin{abstract}

Maternal and child mortality is a public health problem that disproportionately affects low- and middle-income countries. Every day, 800 women and 6,700 newborns die from complications related to pregnancy or childbirth. And for every maternal death, about 20 women suffer serious birth injuries. However, nearly all of these deaths and 
negative health outcomes
are preventable. Midwives are key to revert this situation, and thus it is essential to strengthen their capacities and the quality of their education. This is the aim of the Safe Delivery App, a digital job aid and learning tool to enhance the knowledge, 
confidence and skills
of health practitioners. Here, we use the behavioral logs of the App to implement a recommendation system that presents each midwife with suitable contents to continue gaining 
expertise. We focus on predicting the click-through rate, the probability that a given user will click on a recommended content. We evaluate four deep learning models and show that all of them produce highly accurate~predictions.

\end{abstract}

\section{Introduction}
\label{sec:intro}
Midwives have a profound impact on public health~\citep{Nove2021, UNFPA2020, OConnor2018}. Most maternal and 
newborn
deaths, and other 
negative health outcomes,
occur in low- and middle-income countries and are preventable through timely management by a skilled health professional~\citep{who2019trends}. \hbox{Access} to
quality
midwifery could also improve the long-term health and welfare of mothers and babies~\citep{world2016midwives}.

Precision public health~\citep{dowell2016four} seeks to prevent diseases, promote good habits, and reduce health inequalities by leveraging data and emerging technologies to develop personalized interventions and policies~\citep{buckeridge2020precision, dolley2018big}. The data can come from mobile devices and includes information about diseases, geographical location, behavior or susceptibility at the individual level. However, this approach had not been applied yet to midwifery.

Maternity Foundation~\citep{lund2016association} aims to empower skilled birth attendants to provide a safer birth for mothers and newborns. To that end, they developed the Safe Delivery App~\citep{SDA}, a digital learning tool to enhance the capacity of health practitioners. The App contents are organized into modules that address the main interventions to handle childbirth emergencies, as well as preventative procedures. Modules include videos, action cards, drug lists, practical procedures, and e-learning components.

Applying data science to behavioral logs from the Safe Delivery App may help midwives improve their learning competencies~\citep{guitart2021}. In particular, it could serve to offer each midwife personalized content recommendations, with the ultimate goal of helping them to better assist mothers and newborns. Recommendation systems have been used in many contexts~\citep{itemPrediction2018}, including healthcare applications~\citep{ferretto2017recommender}, where the most typical approaches are collaborative filtering or content-based filtering. 

Here we explore such a recommendation system, focusing instead on the click-through rate (CTR) prediction, namely on estimating each user's probability of clicking on a certain App content. Once identified, the content that each user is more likely to check in the near future can be made more accessible. This should improve the user experience and accelerate the acquisition of critical skills.

\label{sec:modeling}

\section{Click-through rate prediction algorithms for recommendation}

Our strategy involves predictive modeling to foresee a user's response to potential recommendations. For a certain content, the CTR is the rate at which a user presented with the opportunity to consume it will do so. We focus on predicting if a user will check some specific content the next day, a problem to which we can apply advanced CTR prediction models. The main challenges of these models are dealing with high-dimensional, very sparse feature spaces and capturing feature interactions.

Next, we briefly present the models compared in this work. Details concerning the implementation can be found in~\citep{shen2017deepctr}. In all cases, categorical features are one-hot encoded, and features are followed by an embedding layer. The various models differ on the architecture connected to the embedding layer that yields the predicted CTR. They all use some combination of deep neural networks (DNNs, which naturally account for high-order bit-wise feature interactions) and factorization machine (FM) structures (which can efficiently learn low-order vector-wise interactions).

\subsection{Product-based neural networks}

Product-based neural networks (PNNs) were introduced in~\citep{qu2016product} as a deep learning approach to CTR prediction that uses product layers to explore feature interactions. A product layer is built through the pair-wise product of the embedding vectors, and then both the single embedded features and their products are used as the inputs to a multilayer perceptron (MLP).

\subsection{Deep factorization machines}

A deep factorization machine (DeepFM)~\citep{guo2017deepfm} is a neural network architecture that integrates FMs \citep{rendle2010} 
and an MLP. Both elements are directly connected to the embedding layer, and their outputs are fed to the final output layer that yields the CTR prediction. Thus, the main difference between PNNs and DeepFMs is that the feature vector interaction layer is connected to the DNN in the former, and directly to the output in the latter.

\subsection{Extreme deep factorization machines}

An extreme deep factorization machine (xDeepFM)~\citep{lian2018xdeepfm} has a similar architecture to that of the DeepFM, but with the FM replaced by a compressed interaction network (CIN), which is an extension of the crossover network~\citep{wang2017deep}. It receives as input a matrix with each row corresponding to the embedding of a feature, and each of its successive layers outputs a matrix representing interactions between the initial input matrix and the previous layer. This allows the model to measure explicitly high-order vector-wise feature interactions in a scalable manner. Each hidden layer is connected to the output, by first sum pooling on each layer and then concatenating the pooling vectors.

\subsection{Dual input-aware factorization machines}

The dual input-aware factorization machine (DIFM) was introduced in~\citep{lu2020dual}, building on the success of other FM extensions~\citep{juan2016field, yu2019}. All these extensions try to overcome the main limitation of \hbox{FM-based} models, which assign the same weight to a given feature in all possible feature interaction combinations. In the case of DIFMs, the DNN used in~\citep{yu2019} is replaced with a dual layer that contains an MLP-type structure to consider bit-wise interactions, and also a transformer-type architecture~\citep{vaswani2019} that deals with vector-wise interactions when learning the input-aware factors used for reweighting. These are followed by combination, reweighing and prediction layers.

\begin{figure*}
 \centering
 \includegraphics[height=0.24\textheight]{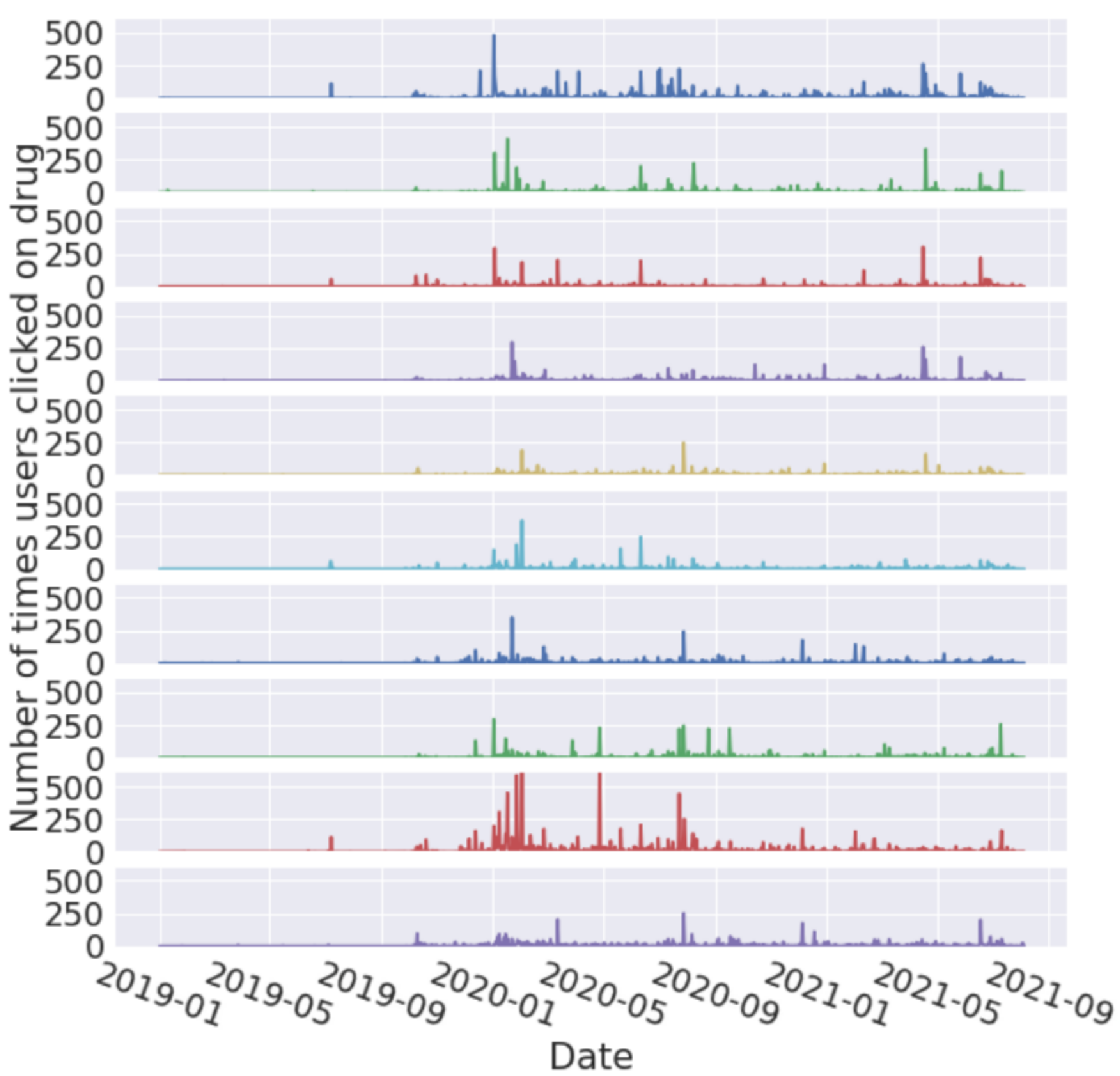}
 \includegraphics[height=0.24\textheight]{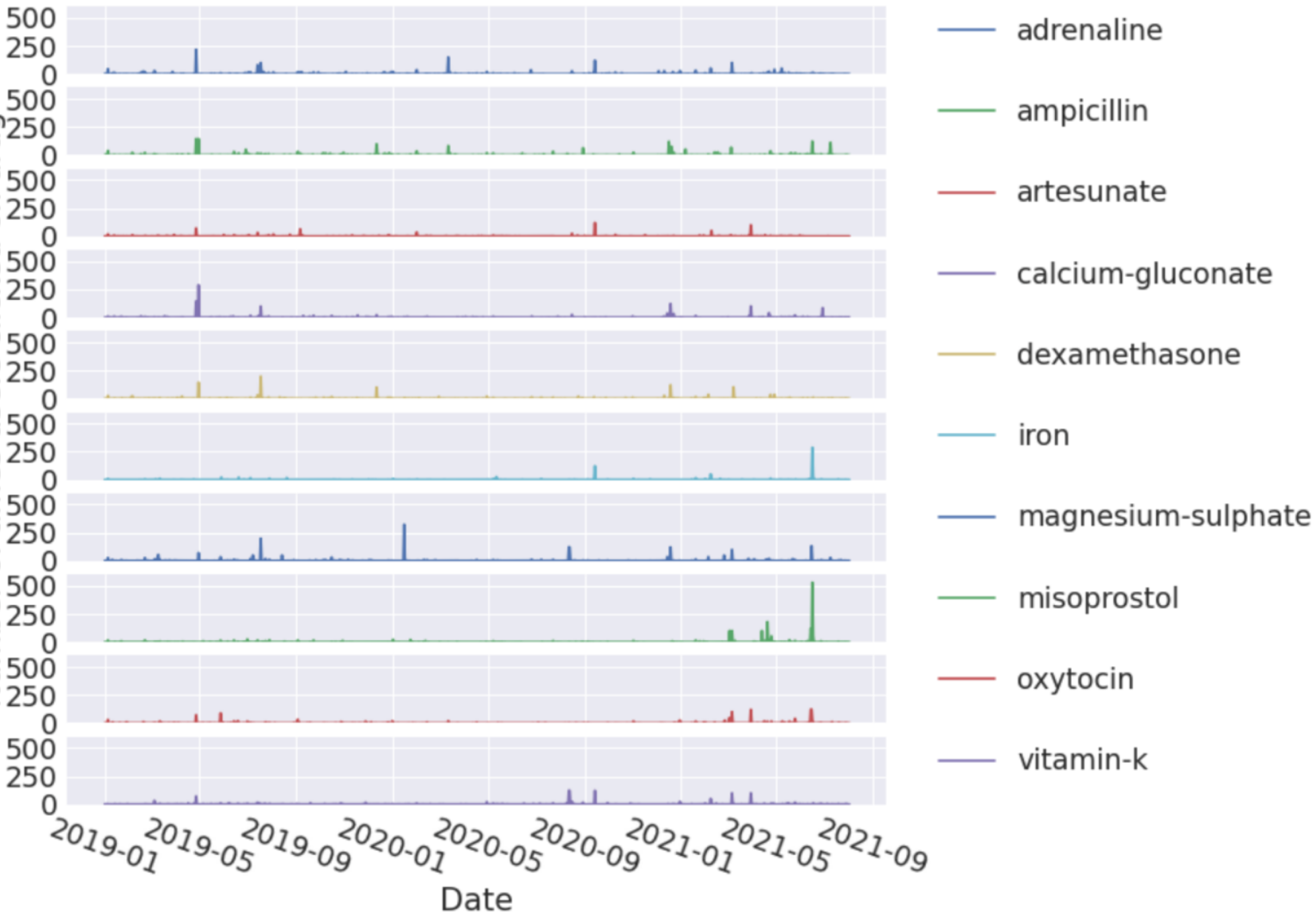}
 \caption{Drug clicks. Total number of daily clicks by Ethiopia's (\emph{left}) and India's (\emph{right}) health professionals on 10 selected drugs covered in the Safe Delivery App.}
 \label{fig:drug_ts}
\end{figure*}

\section{Modeling}

\subsection{Dataset}

The dataset was extracted from the Maternity Foundation’s Safe Delivery App~\citep{SDA} and consisted of user behavioral logs collected between 2019-01-01 and 2021-09-01 (which included 21296 users from India and 1486 from Ethiopia). Each learning \emph{module} from the App includes \emph{videos}, organized in different \emph{chapters}. The App also features a \emph{drugs} section, which lists the most relevant medications for midwives. Those can be grouped in different \emph{drug families}, such as antibiotics, analgesics or antihypertensives. Clicking on any medication displays all the information that health workers need to correctly administer and prescribe it.

User behavioral logs contain a large amount of implicit feedback information on page views, clicks or sequences of checked items. We propose a personalized recommendation algorithm based on the predicted next-day user clicks on four types of contents: video chapters, drugs, video modules 
(i.e., the sets of videos belonging to the same module) 
and drug families. We constructed features related to the user (e.g. connection frequency), the content itself, or both (e.g. the total number of clicks on a certain content), and included external variables such as `day' and `month'. Figure~\ref{fig:drug_ts} depicts real data on the total number of daily clicks for a number of selected drugs. (The corresponding display for video modules is shown in Figures~\ref{fig:video_ethiopia} and \ref{fig:video_india}.)

\subsection{Model specification}

All historical records prior to 2021-03-01 were considered in the training. The dataset was randomly split into training (80\% of the data) and validation (20\%) sets for parameter-tuning purposes. To test the model, we used the 2021-03-01 records. The same process could be used in operational settings.

Models were trained for 10, 20 or 35 epochs (depending on the dataset), using Adam optimization, the ReLu activation function and a batch size of 256. They consisted of 256, 128 and 64 hidden units in three layers. Additionally, the xDeepFM and DIFM models were trained with a 0.5 dropout rate. The latter used tanh instead of ReLu as the activation function and an attention head size of 32.

\section{Results}

The models predict the probability that a given user clicks on each specific content (either an individual drug, a drug family, a module video or one of its chapters). To evaluate model performance, we treated such prediction as a binary classification problem and calculated the area under the curve (AUC)~\citep{hanley1982meaning}. The results are shown in Table~\ref{tab:AUC}. 
Additionally, Table~\ref{tab:RMSE} lists the root-mean-square error (RMSE) across users and contents; see e.g. \citep{Botchkarev2019}. 
This gives an idea of how close to 0 and 1 the predicted CTR probabilities were for negative and positive occurrences, respectively. 

\begin{table*}\centering
  \caption{Area under the ROC curve (AUC) for each model (PNN, DeepFM, xDeepFM and DIFM) and each of the recommender targets (drug, drug family, video chapter and video module). The values were calculated using datasets from Ethiopia and India.}
  \label{tab:AUC}
  \begin{tabular}{llcccc}
    \toprule
    Dataset & Country & \multicolumn{4}{c}{Model}\\
    \cmidrule(lr){3-6}
    & & PNN & DeepFM & xDeepFM & DIFM  \\
    \midrule
    \texttt{Drug} 
      & \texttt{Ethiopia} & 0.9519 & 0.9684 & 0.9674 & 0.9766 \\
      & \texttt{India}    & 0.9694 & 0.9739 & 0.9587 & 0.9735 \\
    \texttt{Drug family} 
      & \texttt{Ethiopia} & 0.9853 & 0.9310 & 0.9696 & 0.9700 \\
      & \texttt{India}    & 0.9664 & 0.9717 & 0.9645 & 0.9742 \\
    \texttt{Video chapter} 
      & \texttt{Ethiopia} & 0.8727 & 0.8814 & 0.9070 & 0.8199 \\
      & \texttt{India}    & 0.9654 & 0.9712 & 0.9692 & 0.9363 \\
    \texttt{Video module}
      & \texttt{Ethiopia}  & 0.9074 & 0.9019 & 0.9055 & 0.8653 \\
      & \texttt{India}     & 0.9657 & 0.9752 & 0.9727 & 0.9319 \\
    \bottomrule
  \end{tabular}
\end{table*}

All models performed very well for all types of content, with AUC values above 0.8 (and above 0.9 almost in all instances) and low RMSE (which in this case also ranges between 0 and 1).

As expected, predictions are better for drug families than for individual drugs. However, the difference is not significant. This means our models are able to foresee not only the type of health problem that will be encountered, but also the specific drugs that will be checked, which apparently suggests that midwives always tend to use the drugs they are already familiar with.

The smaller training-sample sizes in the Ethiopian case foreseeably translated into less predictive power. However, the results are still very accurate, which leads us to conclude that our models can effectively learn from relatively small samples.

The fact that the xDeepFM and DIFM models did not perform significantly better than the \hbox{simpler} PNN and DeepFM ones suggests that high-order vector-wise feature interactions are not relevant in this problem for the selected features. The DIFM model produces the best AUC values when forecasting drug clicks, while consistently yielding the worst results (though not by much) for video prediction. This could mean that input-aware reweighting does have some importance in the drug~case. 

However, the DIFM model is always among the worst choices in terms of the RMSE; see Table~\ref{tab:RMSE}. On the other hand, the DeepFM and xDeepFM errors are very similar and the lowest ones for all content types. This means they should perform better than the other two models, provided we tune the probability threshold above which we consider there is a predicted positive occurrence. The DeepFM architecture is simpler and can be trained more efficiently, and that makes it the preferred candidate to be implemented in an operational setting.

This is further supported by Figures~\ref{fig:rmse_drug_ethiopia} and \ref{fig:rmse_drug_india}, which show the RMSE of the individual-drug predictions produced by each model for Ethiopia and India, respectively. These figures reveal that the error is far from being uniformly distributed, with the largest contributions coming from only a few drugs. They show, too, that there is a group of drugs for which all models present large errors (with the PNN and DIFM models exhibiting additional peaks). 

\section{Summary and conclusions}

We have shown that state-of-the-art CTR prediction models can successfully foretell whether a user will check a certain video or drug content within the next day. Therefore, they can be used to make that content more accessible to the user. This could facilitate midwives' everyday practice and skill acquisition, ultimately resulting in a better care for the mothers and babies they are assisting.

While all tested models yielded highly accurate predictions, our results suggest that the simpler DeepFM architecture would suffice to address this content recommendation problem in a production environment. This seems to indicate that explicitly considering high-order vector-wise feature interactions and feature-aware reweighting has no significant effect in this particular problem.

On the whole, in terms of overall performance and efficiency, the DeepFM model appears like the ideal candidate to be used in production. 

\section*{Acknowledgements} 
All the data used in this analysis comes from the Safe Delivery App logs and belongs to Maternity Foundation. For inquiries regarding its use, please contact them at mail@maternity.dk. 
The authors thank Javier Grande for his careful review of the manuscript.

\bibliographystyle{plain} 
\bibliography{main}

\begin{thebibliography}{25}
\providecommand{\natexlab}[1]{#1}
\providecommand{\url}[1]{\texttt{#1}}
\expandafter\ifx\csname urlstyle\endcsname\relax
  \providecommand{\doi}[1]{doi: #1}\else
  \providecommand{\doi}{doi: \begingroup \urlstyle{rm}\Url}\fi

\bibitem[Bertens et~al.(2018)Bertens, Guitart, Chen, and
  Peri{\'a}{\~n}ez]{itemPrediction2018}
Paul Bertens, Anna Guitart, Pei~Pei Chen, and {\'A}frica Peri{\'a}{\~n}ez.
\newblock A machine-learning item recommendation system for video games.
\newblock In \emph{2018 IEEE Conference on Computational Intelligence and Games
  (CIG)}, pages 1--4, Maastrich, Netherlands, 2018. IEEE.

\bibitem[Botchkarev(2019)]{Botchkarev2019}
Alexei Botchkarev.
\newblock A new typology design of performance metrics to measure errors in
  machine learning regression algorithms.
\newblock \emph{Interdisciplinary Journal of Information, Knowledge, and
  Management}, 14:\penalty0 045--076, 2019.
\newblock ISSN 1555-1237.
\newblock \doi{10.28945/4184}.
\newblock URL \url{http://dx.doi.org/10.28945/4184}.

\bibitem[Buckeridge(2020)]{buckeridge2020precision}
David~L Buckeridge.
\newblock Precision, equity, and public health and epidemiology informatics--a
  scoping review.
\newblock \emph{Yearbook of Medical Informatics}, 29\penalty0 (01):\penalty0
  226--230, 2020.

\bibitem[Dolley(2018)]{dolley2018big}
Shawn Dolley.
\newblock Big data’s role in precision public health.
\newblock \emph{Frontiers in public health}, 6:\penalty0 68, 2018.

\bibitem[Dowell et~al.(2016)Dowell, Blazes, and
  Desmond-Hellmann]{dowell2016four}
Scott~F Dowell, David Blazes, and Susan Desmond-Hellmann.
\newblock Four steps to precision public health.
\newblock \emph{Nature News}, 540\penalty0 (7632):\penalty0 189, 2016.

\bibitem[Ferretto et~al.(2017)Ferretto, Cervi, and
  de~Marchi]{ferretto2017recommender}
Luciano~Rodrigo Ferretto, Cristiano~Roberto Cervi, and Ana Carolina~Bertoletti
  de~Marchi.
\newblock Recommender systems in mobile apps for health a systematic review.
\newblock In \emph{2017 12th Iberian Conference on Information Systems and
  Technologies (CISTI)}, pages 1--6. IEEE, 2017.

\bibitem[Foundation(2021)]{SDA}
Maternity Foundation.
\newblock Safe delivery app, 2021.
\newblock Accessed: 2021-09-03.

\bibitem[Fund(2020)]{UNFPA2020}
United Nations~Population Fund.
\newblock Cost of ending preventable maternal deaths.
\newblock In \emph{Costing the three transformative results}, pages 11--17,
  January 2020.
\newblock URL
  \url{{https://www.unfpa.org/sites/default/files/pub-pdf/Transformative_results_journal_23-online.pdf}}.

\bibitem[Guitart et~al.(2021)Guitart, Fern{\'a}ndez~del R{\'i}o,
  Peri{\'a}{\~n}ez, and Bellhouse]{guitart2021}
Anna Guitart, Ana Fern{\'a}ndez~del R{\'i}o, {\'A}frica Peri{\'a}{\~n}ez, and
  Lauren Bellhouse.
\newblock Midwifery learning and forecasting: Predicting content demand with
  user-generated logs, 2021.

\bibitem[Guo et~al.(2017)Guo, Tang, Ye, Li, and He]{guo2017deepfm}
Huifeng Guo, Ruiming Tang, Yunming Ye, Zhenguo Li, and Xiuqiang He.
\newblock Deepfm: a factorization-machine based neural network for ctr
  prediction, 2017.

\bibitem[Hanley and McNeil(1982)]{hanley1982meaning}
James~A Hanley and Barbara~J McNeil.
\newblock The meaning and use of the area under a receiver operating
  characteristic (roc) curve.
\newblock \emph{Radiology}, 143\penalty0 (1):\penalty0 29--36, 1982.

\bibitem[Juan et~al.(2016)Juan, Zhuang, Chin, and Lin]{juan2016field}
Yuchin Juan, Yong Zhuang, Wei-Sheng Chin, and Chih-Jen Lin.
\newblock Field-aware factorization machines for ctr prediction.
\newblock In \emph{Proceedings of the 10th ACM conference on recommender
  systems}, pages 43--50, 2016.

\bibitem[Lian et~al.(2018)Lian, Zhou, Zhang, Chen, Xie, and
  Sun]{lian2018xdeepfm}
Jianxun Lian, Xiaohuan Zhou, Fuzheng Zhang, Zhongxia Chen, Xing Xie, and
  Guangzhong Sun.
\newblock xdeepfm: Combining explicit and implicit feature interactions for
  recommender systems.
\newblock In \emph{Proceedings of the 24th ACM SIGKDD International Conference
  on Knowledge Discovery and Data Mining}, pages 1754--1763, 2018.

\bibitem[Lu et~al.(2020)Lu, Yu, Chang, Wang, Li, and Yuan]{lu2020dual}
Wantong Lu, Yantao Yu, Yongzhe Chang, Zhen Wang, Chenhui Li, and Bo~Yuan.
\newblock A dual input-aware factorization machine for ctr prediction.
\newblock In \emph{IJCAI}, pages 3139--3145, 2020.

\bibitem[Lund et~al.(2016)Lund, Boas, Bedesa, Fekede, Nielsen, and
  S{\o}rensen]{lund2016association}
Stine Lund, Ida~Marie Boas, Tariku Bedesa, Wondewossen Fekede, Henriette~Svarre
  Nielsen, and Bjarke~Lund S{\o}rensen.
\newblock Association between the safe delivery app and quality of care and
  perinatal survival in ethiopia: a randomized clinical trial.
\newblock \emph{JAMA pediatrics}, 170\penalty0 (8):\penalty0 765--771, 2016.

\bibitem[Nove et~al.(2021)Nove, Friberg, de~Bernis, McConville, Moran,
  Najjemba, ten Hoope-Bender, Tracy, and Homer]{Nove2021}
Andrea Nove, Ingrid~K Friberg, Luc de~Bernis, Fran McConville, Allisyn~C Moran,
  Maria Najjemba, Petra ten Hoope-Bender, Sally Tracy, and Caroline~SE Homer.
\newblock Potential impact of midwives in preventing and educing maternal and
  neonatal mortality and stillbirths: a lives saved tool modelling study.
\newblock \emph{The Lancet Global Health}, 9\penalty0 (1):\penalty0 24--32,
  2021.

\bibitem[O'connor(2018)]{OConnor2018}
Siobhan O'connor.
\newblock Big data and data science in health care: What nurses and midwives
  need to know, 2018.

\bibitem[Organization et~al.(2016)]{world2016midwives}
World~Health Organization et~al.
\newblock Midwives voices, midwives realities. findings from a global
  consultation on providing quality midwifery care, 2016.

\bibitem[Qu et~al.(2016)Qu, Cai, Ren, Zhang, Yu, Wen, and Wang]{qu2016product}
Yanru Qu, Han Cai, Kan Ren, Weinan Zhang, Yong Yu, Ying Wen, and Jun Wang.
\newblock Product-based neural networks for user response prediction.
\newblock In \emph{2016 IEEE 16th International Conference on Data Mining
  (ICDM)}, pages 1149--1154. IEEE, 2016.

\bibitem[Rendle(2010)]{rendle2010}
Steffen Rendle.
\newblock Factorization machines.
\newblock In \emph{2010 IEEE International conference on data mining}, pages
  995--1000. IEEE, 2010.

\bibitem[Shen(2017)]{shen2017deepctr}
Weichen Shen.
\newblock Deepctr: Easy-to-use,modular and extendible package of deep-learning
  based ctr models.
\newblock \url{https://github.com/shenweichen/deepctr}, 2017.

\bibitem[Vaswani et~al.(2017)Vaswani, Shazeer, Parmar, Uszkoreit, Jones, Gomez,
  Kaiser, and Polosukhin]{vaswani2019}
Ashish Vaswani, Noam Shazeer, Niki Parmar, Jakob Uszkoreit, Llion Jones,
  Aidan~N. Gomez, \L{}ukasz Kaiser, and Illia Polosukhin.
\newblock Attention is all you need.
\newblock In \emph{Proceedings of the 31st International Conference on Neural
  Information Processing Systems}, NIPS'17, page 6000–6010, Red Hook, NY,
  USA, 2017. Curran Associates Inc.
\newblock ISBN 9781510860964.

\bibitem[Wang et~al.(2017)Wang, Fu, Fu, and Wang]{wang2017deep}
Ruoxi Wang, Bin Fu, Gang Fu, and Mingliang Wang.
\newblock Deep \& cross network for ad click predictions.
\newblock In \emph{Proceedings of AdKDD and TargetAd}, 2017.

\bibitem[World Health~Organization and Bank(2019)]{who2019trends}
United Nations Population~Fund World Health~Organization, UNICEF and The~World
  Bank.
\newblock Trends in maternal mortality: 2000 to 2017: {E}stimates by {WHO},
  {UNICEF}, 2019.

\bibitem[Yu et~al.(2019)Yu, Wang, and Yuan]{yu2019}
Yantao Yu, Zhen Wang, and Bo~Yuan.
\newblock An input-aware factorization machine for sparse prediction.
\newblock In \emph{Proceedings of the Twenty-Eighth International Joint
  Conference on Artificial Intelligence, {IJCAI-19}}, pages 1466--1472.
  International Joint Conferences on Artificial Intelligence Organization, 7
  2019.
\newblock \doi{10.24963/ijcai.2019/203}.
\newblock URL \url{https://doi.org/10.24963/ijcai.2019/203}.

\end{thebibliography}

\appendix
\counterwithin{figure}{section}
\counterwithin{table}{section}

\newpage
\section{Additional figures and tables}\label{annex}

\begin{table*}[h!]\centering
  \caption{Root-mean-square error (RMSE) for each model (PNN, DeepFM, xDeepFM and DIFM) and each of the recommender targets (drug, drug family, video chapter and video module). The values were calculated using datasets from Ethiopia and India.}
  \label{tab:RMSE}
  \begin{tabular}{llcccc}
    \toprule
    Dataset & Country & \multicolumn{4}{c}{Model}\\
    \cmidrule(lr){3-6}
    & & PNN & DeepFM & xDeepFM & DIFM  \\
    \midrule
    \texttt{Drug} 
      & \texttt{Ethiopia} & 0.0886 & 0.0824 & 0.0832 & 0.1089 \\
      & \texttt{India}    & 0.0330 & 0.0263 & 0.0256 & 0.0404 \\
    \texttt{Drug Family} 
      & \texttt{Ethiopia} & 0.1219 & 0.1216 & 0.1205 & 0.1362 \\
      & \texttt{India}    & 0.0487 & 0.0376 & 0.0373 & 0.0586 \\
    \texttt{Video Chapter} 
      & \texttt{Ethiopia} & 0.1587 & 0.1029 & 0.1042 & 0.1289 \\
      & \texttt{India}    & 0.1375 & 0.0555 & 0.0553 & 0.1113 \\
    \texttt{Video Module} 
      & \texttt{Ethiopia} & 0.1857 & 0.1789 & 0.1789 & 0.2504 \\
      & \texttt{India}    & 0.1055 & 0.0929 & 0.0939 & 0.1955 \\
    \bottomrule
  \end{tabular}
\end{table*}
\clearpage

 \begin{figure*}
     \centering
     \includegraphics[width=1\textwidth]{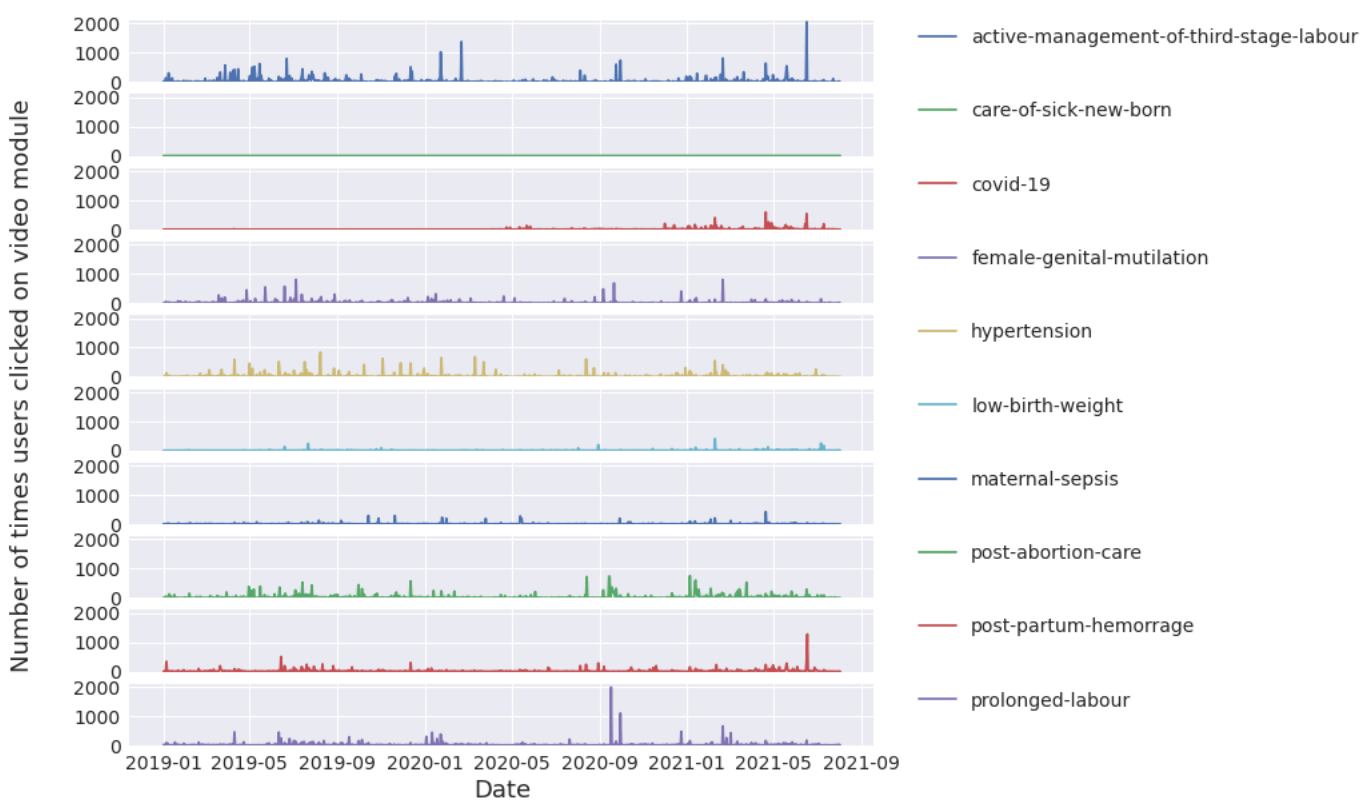}
     \caption{Video module clicks. Total number of daily clicks by Ethiopia's health professionals on 10 selected video modules from the Safe Delivery App.}
     \label{fig:video_ethiopia}
 \end{figure*}

 \begin{figure*}
     \centering
     \includegraphics[width=1\textwidth]{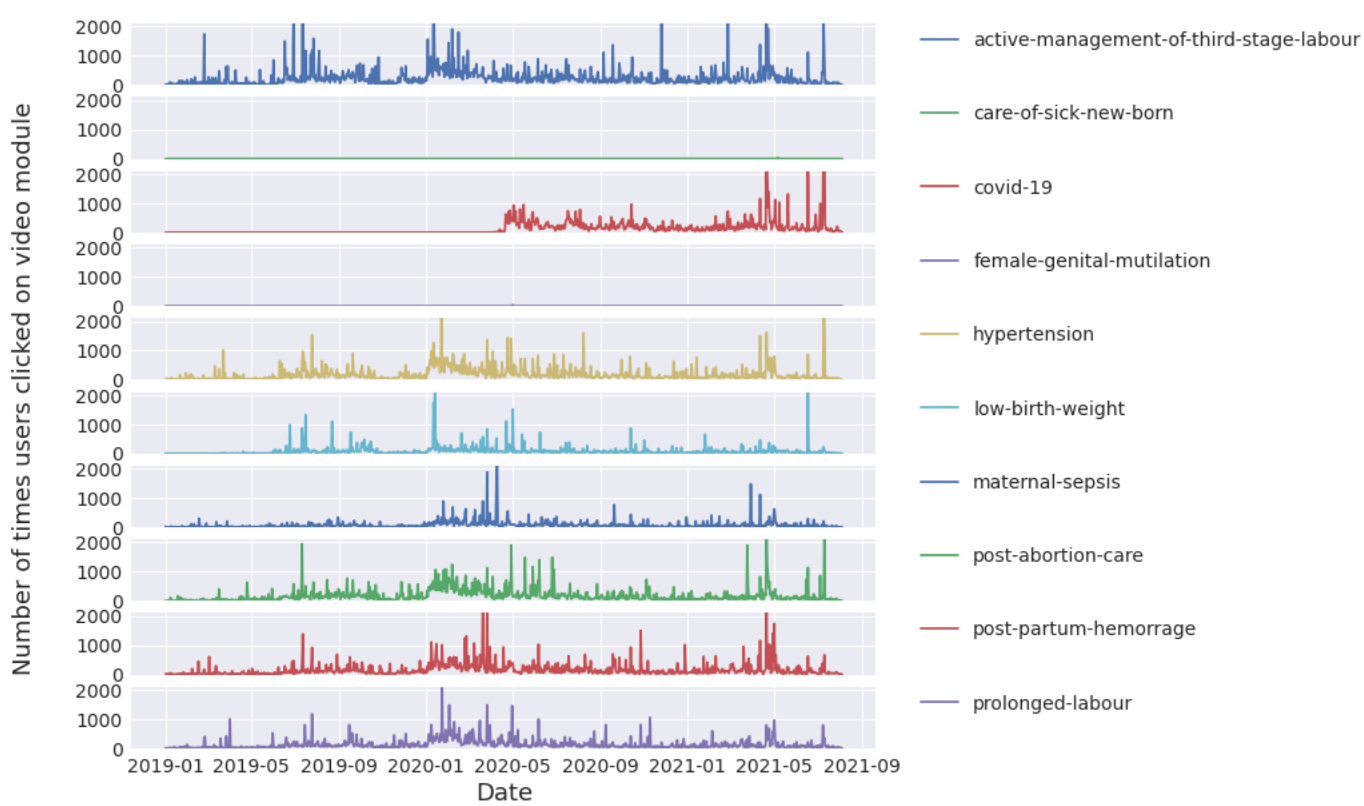}
     \caption{Video module clicks. Total number of daily clicks by India's health professionals on 10 selected video modules from the Safe Delivery App.}
     \label{fig:video_india}
 \end{figure*}

 \begin{figure*}
     \centering
     \includegraphics[width=1\textwidth]{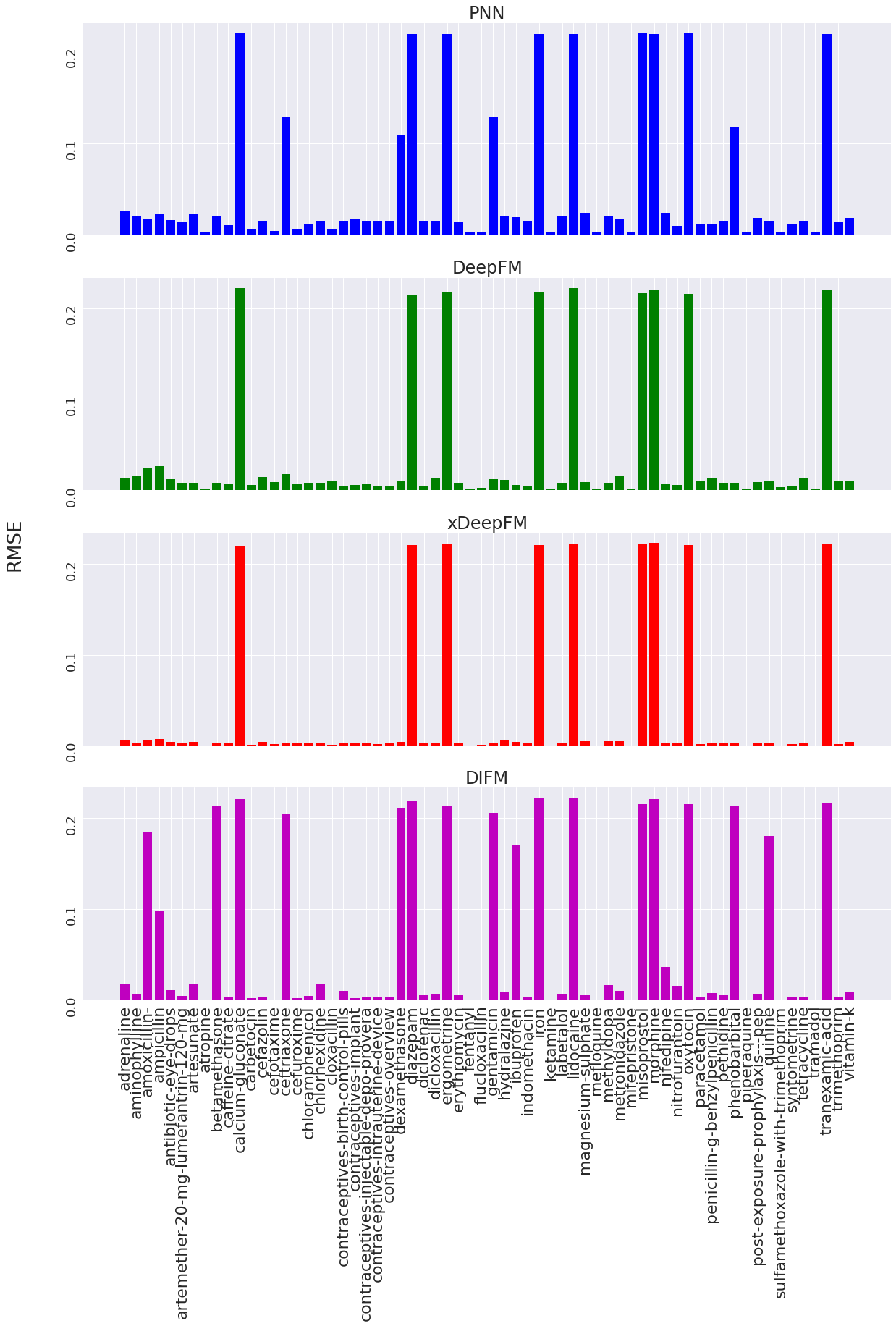}
     \caption{Root-mean-square error (RMSE) per drug for the predictions of the PNN, DeepFM, xDeepFM and DIFM models across all users in Ethiopia.}
     \label{fig:rmse_drug_ethiopia}
 \end{figure*}
 
  \begin{figure*}
     \centering
     \includegraphics[width=1\textwidth]{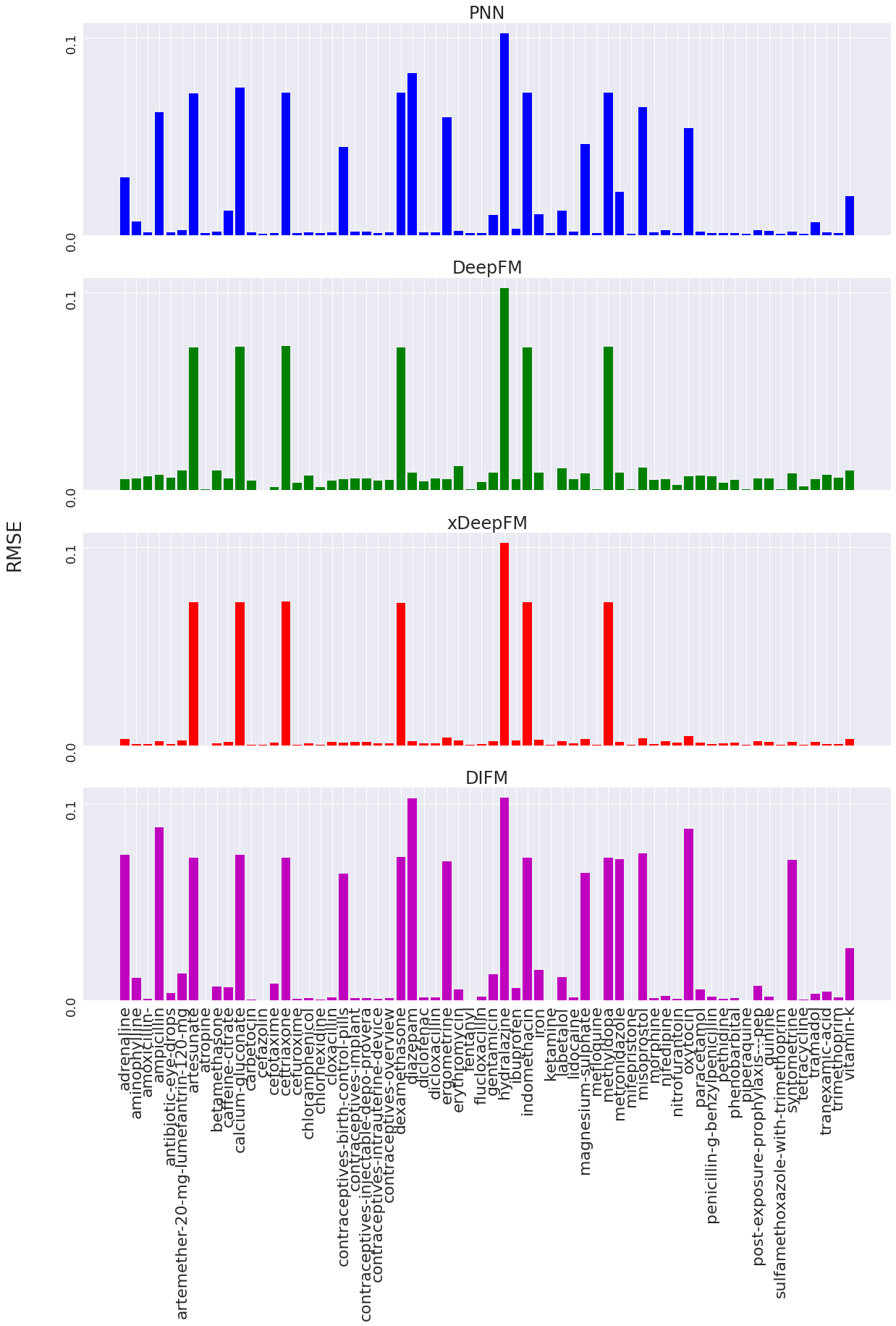}
     \caption{Root-mean-square error (RMSE) per drug for the predictions of the PNN, DeepFM, xDeepFM and DIFM models across all users in India.}
     \label{fig:rmse_drug_india}
 \end{figure*}

\end{document}